\documentclass[letterpaper, 10 pt, conference]{ieeeconf}  

\IEEEoverridecommandlockouts                              

\usepackage{graphics} 
\usepackage{epsfig} 
\usepackage{color}
\usepackage{threeparttable}
\usepackage{amsmath,amssymb}
\usepackage{mathrsfs}
\usepackage{multicol}
\usepackage{multirow}
\usepackage{marvosym}

\usepackage{amsthm}

\title{\LARGE \bf
 Planning Assembly Sequence with Graph Transformer
}

\author{Lin Ma$^{1}$, Jiangtao Gong$^{1}$\textsuperscript{\Letter}, Hao Xu$^{2}$, Hao Chen$^{2}$, Hao Zhao$^{1}$, Wenbing Huang$^{1}$ and Guyue Zhou$^{1}$
\thanks{$^{1}$Institute for AI Industry Research (AIR), Tsinghua University, 10080, Haidian District, Beijing, P.R.China.
        {\tt\small lastnamefirstname@air.tsinghua.edu.cn}}%
\thanks{$^{2}$ Qianzhi Technology, China
        {\tt\small hao.xu.chn@gmail.com}}%
}

\begin{document}

\maketitle
\thispagestyle{empty}
\pagestyle{empty}

\begin{abstract}
Assembly sequence planning (ASP) is the essential process for modern manufacturing, proven to be NP-complete thus its effective and efficient solution has been a challenge for researchers in the field.
In this paper, we present a graph-transformer based framework for the ASP problem which is trained and demonstrated on a self-collected ASP database.
The ASP database contains a self-collected set of LEGO models. The LEGO model is abstracted to a heterogeneous graph structure after a thorough analysis of the original structure and feature extraction.
The ground truth assembly sequence is first generated by brute-force search and then adjusted manually to in line with human rational habits.
Based on this self-collected ASP dataset, we propose a heterogeneous graph-transformer framework to learn the latent rules for assembly planning.
We evaluated the proposed framework in a series of experiment. The results show that the similarity of the predicted and ground truth sequences can reach 0.44, a medium correlation measured by Kendall's $\tau$.
Meanwhile, we compared the different effects of node features and edge features and generated a feasible and reasonable assembly sequence as a benchmark for further research.
Our data set and code is available on https://github.com/AIR-DISCOVER/ICRA\_ASP.
\end{abstract}

\section{INTRODUCTION}

To facilitate automatic assembly in a wide range of fields, such as the furniture industry, auto manufacturing industry, and arts and crafts industry, assembly sequence planning is a key procedure after product design.

 The dominant methods for ASP problems are assembly-by-disassembly planning and the combinatorial optimization algorithm. To be detailed, the main task of Assembly-by-disassembly planning is finding a feasible removing path by recursively removing parts from the assembly product \cite{dorn2022assembly} \cite{tseng2011green}. Moreover, the combinatorial optimization algorithm aims at determining the precedence relationship among all parts under complex constraints \cite{jimenez2013survey} \cite{ghandi2015breakout} using dynamic programming.

ASP is proven to be NP-complete, so its effective and efficient solution has been a challenge for researchers in the field. Consequently, there is much room for progress in this research. Deep learning methods have been widely used in various areas and proven to be effective, so we expect to use deep learning methods to learn the latent rules of an assembly problem from previous experience. However, the two dominant methods mentioned above can only find a solution for a specific assembly model. As a result, there is no dataset specialized for ASP problems to train a deep learning model.

The datasets most related to our task are illustrated as follows. A main application of ASP is generating step-by-step instructions for furniture and household articles, which are also universally used in many research \cite{agrawala2003designing}.  80 IKEA furniture models are present and applied to assembly environment design in \cite{lee2021ikea}. These furniture models are made up of several parts without complex block relationships, so the main task is to ensure stability or visibility in the assembly process. Moreover, a large-scale part-level 3D dataset containing various objects in indoor scenes was released \cite{mo2019partnet}.  This dataset enables many 3D generative tasks at the part level, such as estimating the location and pose of parts \cite{zhang20223d} and learning a consistent part order for a given object category \cite{wu2020pq}. Except for furniture and household items, 3D shapes simulated by the LEGO model are generated via sequence assembly \cite{thompson2020building} \cite{kim2020combinatorial}. However, the aforementioned assembly datasets are mostly designed or collected for generative tasks rather than sequence assembly planning, so a dataset specifically for ASP problem is demanded. A suitable assembly dataset composed of a moderate number of instances and containing common block relationships would provide us with more opportunities to establish universal solutions to the ASP problem.

However, the collection of assembly instances is no easy task for the following reasons. Firstly, the instances are supposed to be sufficient for training and homogeneous with each other. Nevertheless, the majority of the industries only release a few temples. Secondly, even if we have the temples, the assembly sequences, namely, the ground truth, are always not available.

\begin{figure}
\centering
  \includegraphics[width=1.05\columnwidth]{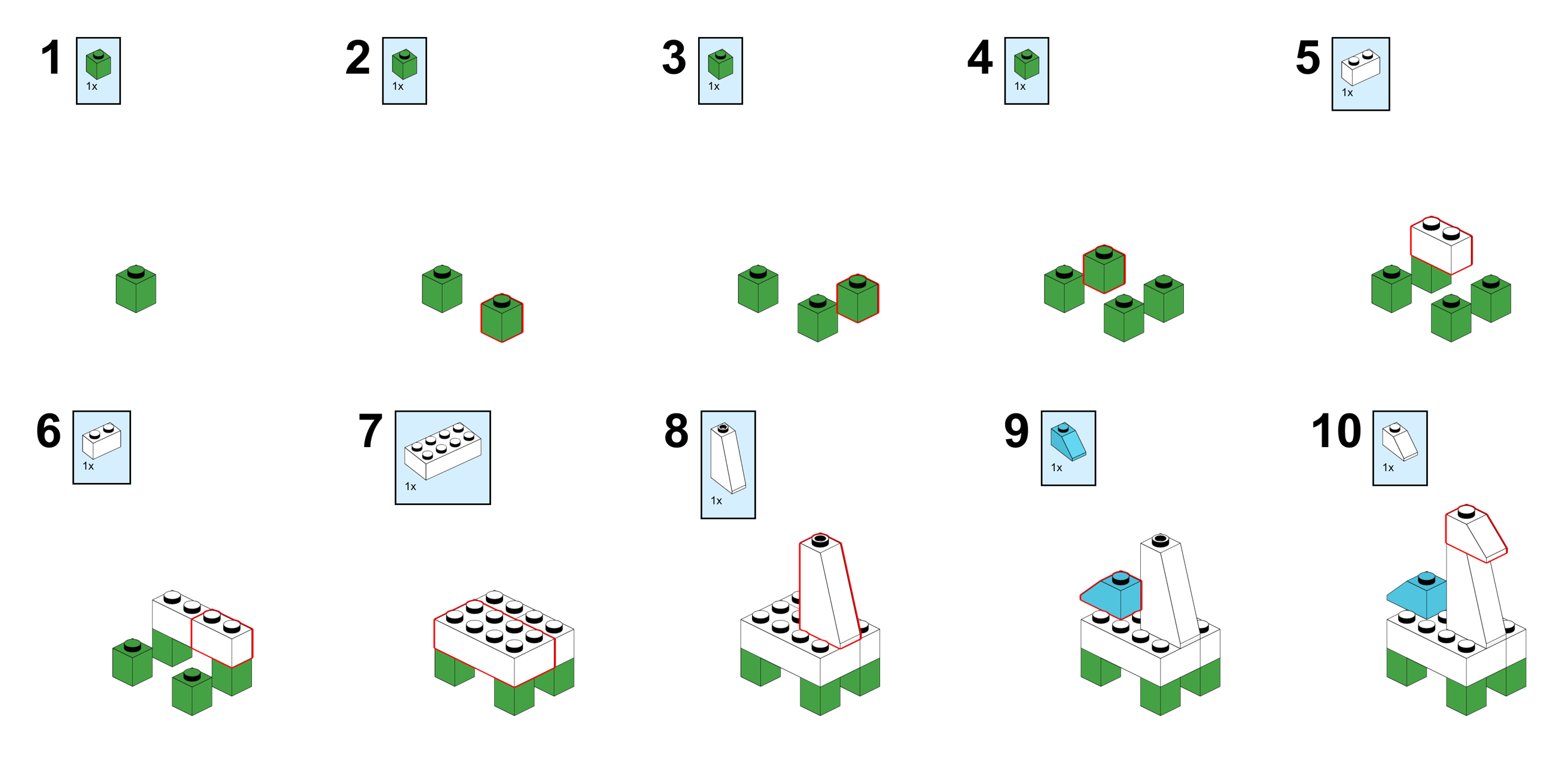}
  \caption{An example of Assembly sequence planning}
  \label{fig:intro}
\end{figure}

In this paper, we present a collective LEGO dataset for the ASP task. Structures built from LEGO are complex enough to approach real-world design and have a large number of models designed and uploaded by users. Unfortunately, not all models are practical, so we made some adjustments, which will be illustrated in section III. Moreover, we propose a graph-transformer-based framework for ASP. Our goal is to train a network to learn the latent rules of the assembly sequence and automatically generate a feasible assembly sequence according to a design template.

The main contribution of this work can be summarized as follows.
\begin{enumerate}

    \item A self-collected ASP dataset using user-defined models from LEGO Studio and feasible assembly sequences by brute force method and manual adjustment as ground truth;
    \item A graph-transformer based framework for ASP problem, with a heterogeneous graph attention network to encoder the models, which are decoded with the attention mechanism to generate assembly sequences;
    \item A series of experiment to evaluate the effectiveness of the proposed framework and offer the benchmark for this ASP problem.
\end{enumerate}

\section{Ralated Work}

\textbf{Assembly sequence planning} ASP problems are usually solved by assembly-by-disassembly planning and the combinatorial optimization algorithm. The assembly-by-disassembly method finds a feasible disassembly order via motion planning \cite{sundaram2001disassembly}.  Being NP-complete, it requires numerous search operations even to find a feasible disassembly path.  Thus researchers compress the search space using a tree structure, such as "Rapidly-exploring random tree" \cite{lavalle1998rapidly} and Expansive Vornoi tree \cite{karaman2011sampling}. Nevertheless, the combinatorial optimization algorithm always implements the heuristic methods, such as Ant Colony optimization algorithm \cite{han2021ant}, Gray Wolf optimization algorithm \cite{ab2017hybrid}, and  Artificial Neural Networks \cite{philip1992design}.

ASP problems have not been extensively studied by deep learning methods. To the best of our knowledge, the only method related to deep learning is proposed in \cite{zhao2019aspw}. We believe there is still much room for progress in this research.

\textbf{LEGO model} Since LEGO models are complex enough to approach real-world design, it is common for researchers and designers to regard the LEGO model as an abstraction of a physical object. By studying the LEGO model, we can increase our cognition of the real world and be inspired to solve practical problems. For example, LEGO models are used to approximate actual 3D shapes while ensuring that the final products have some good properties such as connectivity and stability \cite{xu2020computational} \cite{stephenson2016multi}.  Furthermore, the LEGO models abstracted from physical objects are applied to the deep generative models for 3D shape generative tasks \cite{thompson2020building} \cite{kim2020combinatorial}.

Besides, as an abstraction of real-world objects, the LEGO model is a powerful tool for promoting and testing human cognitive abilities\cite{horikoshi2021teaching}.  Moreover, cognitive loads are usually studied during LEGO assembly tasks \cite{yang2019influences}.

In the above scenario, the LEGO model is abstracted into LEGO Model Representation graph \cite{kim2014survey} \cite{thompson2020building}, with a single vertex in the graph representing a LEGO brick (regular brick, plate, or tile) and an edge representing the linkage between two LEGO bricks. This is consistent with real-world assembly problems. For example, furniture and cars are abstracted to graph structures  \cite{mo2019partnet} \cite{dorn2022assembly}.

\textbf{Graph-transformer} The transformer has became the standard architecture for a wild variety of fields. In particular, its variants have been modified to explore the latent feature of graph-structured data and fulfill different tasks for natural language processing  \cite{cai2020graph}, biological molecular structure \cite{dwivedi2020generalization} \cite{ying2021transformers}, social networks \cite{chandrika2022graph}, and academic network \cite{hu2020heterogeneous}.

In a heterogeneous graph, different types of objects are mapped to different types of nodes, and the same is true for edges \cite{yao2020heterogeneous}  \cite{kumar2021user}. This abstraction is more appropriate for containing more complete information.  Accordingly, LEGO models are similarly devised to heterogeneous graphs in our data set. Different node types represent different LEGO bricks and different edge types for different relations between LEGO bricks.

Almost all of the tasks of the graph-transformer framework are node classification, graph classification, link prediction, and text generation, but barely see its application in sequence planning. To the best of our knowledge, this is the first attempt to utilize graph transformers in an assembly sequence planning task.

\section{LEGO-ASP dataset}
\subsection{Data set summary}
\begin{figure}
\centering
  \includegraphics[width=0.48\columnwidth]{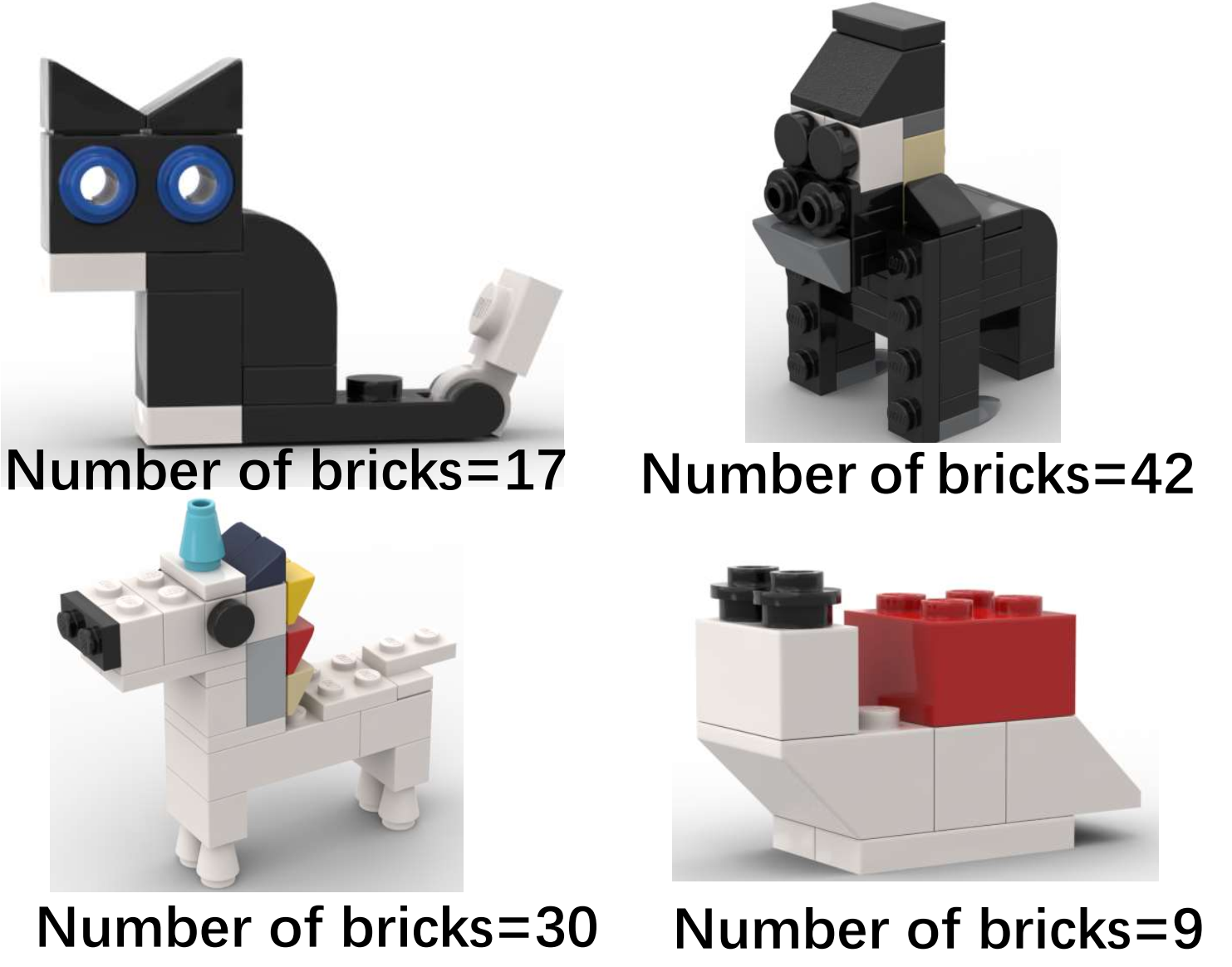}
  \includegraphics[width=0.48\columnwidth]{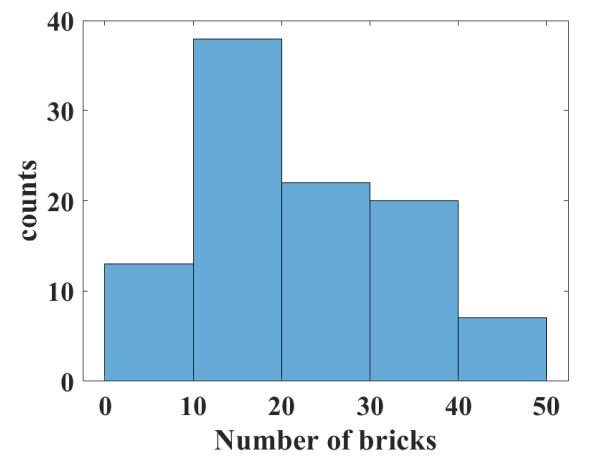}
  \caption{Overview of the LEGO models.}
  \label{fig:LEGO_model}
\end{figure}

We collect 100 LEGO animal models created and uploaded by individual users in LEGO Studio, among which the simplest one is composed of 3 bricks and the most complex one is composed of 44 bricks. The median number of brick numbers in a LEGO model is 19.  As shown in Fig. \ref{fig:LEGO_model}, we present four models with different number of bricks as examples. And the statistics of the number of bricks in a individual LEGO model is demonstrated in the histogram.

\begin{figure}
\centering
  \includegraphics[width=0.48\columnwidth]{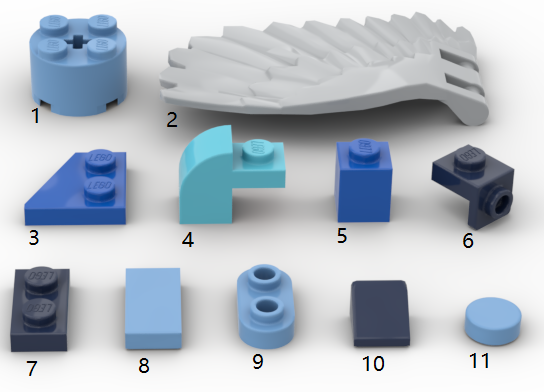}
  \includegraphics[width=0.48\columnwidth]{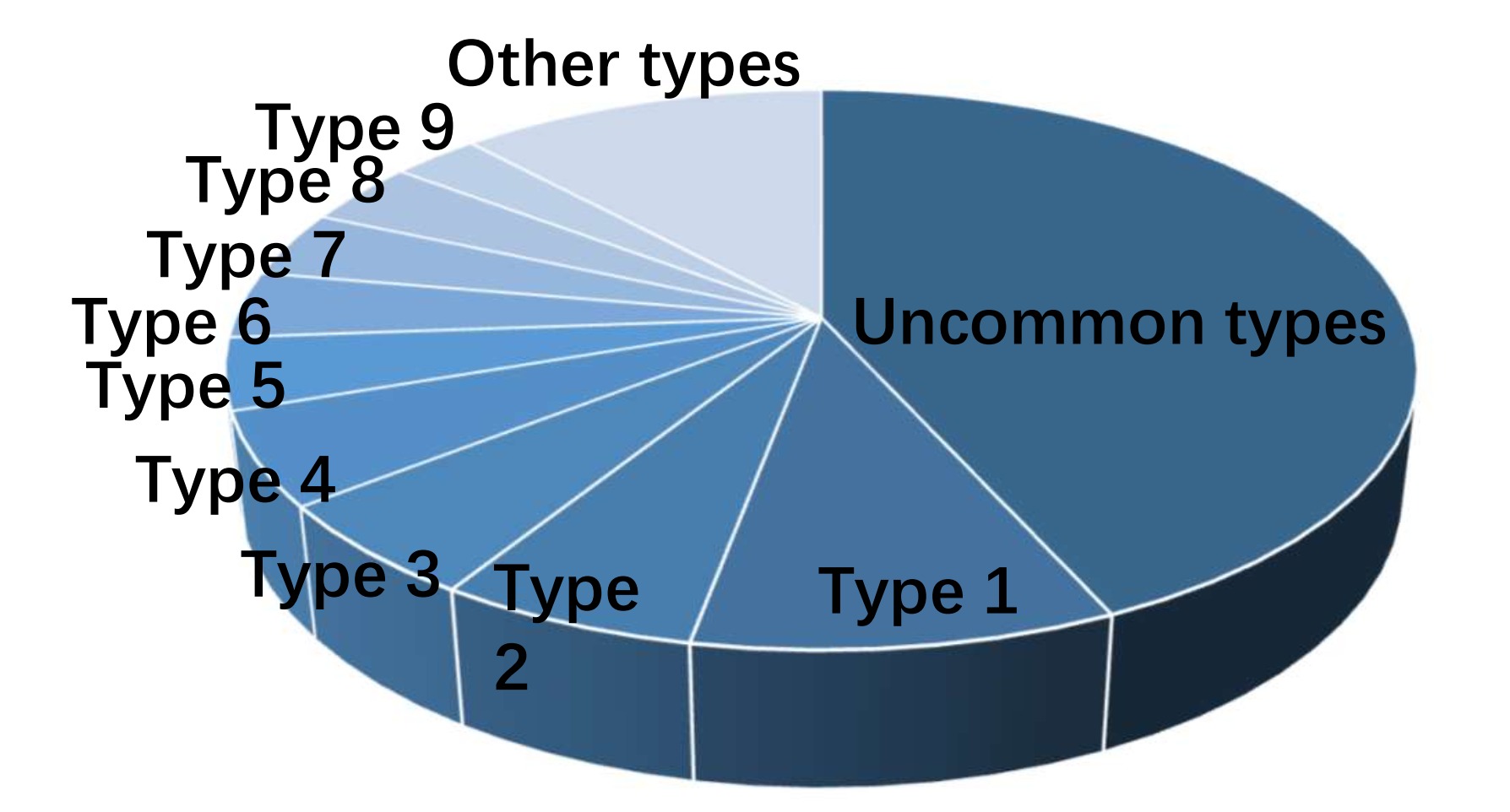}
  \caption{Overview on brick types.}
  \label{fig:brick_type}
\end{figure}

The total number of bricks in all LEGO models is 2127, of which 23 types of bricks are frequently used. The rest are used only a few times or customized, but they take up a non-negligible percentage as a whole. According to the role of bricks, the frequently-used bricks can be roughly categorized into the following three types: regular brick, plate, and tile. In addition, they can also be categorized into four types according to their shape: cubic, curve, slope, and round. At the same time, the bricks are of different sizes. With different combinations of these features, a large variety of bricks exist. We present different types of LEGO bricks in Fig. \ref{fig:brick_type}. For example, the first brick is a regular brick as well as round type.  The second is a customized brick, the seventh is a plate of size 1x2. Moreover, the ratio of different types is presented in the pie chart.

 However, these LEGO models uploaded by individual users may not be carefully checked, so separations and collisions may occur. In this case, we checked all models and fix the separations and collisions, while preserving the original shapes and structures as much as possible.

\begin{figure}
\centering
  \includegraphics[width=0.45\columnwidth]{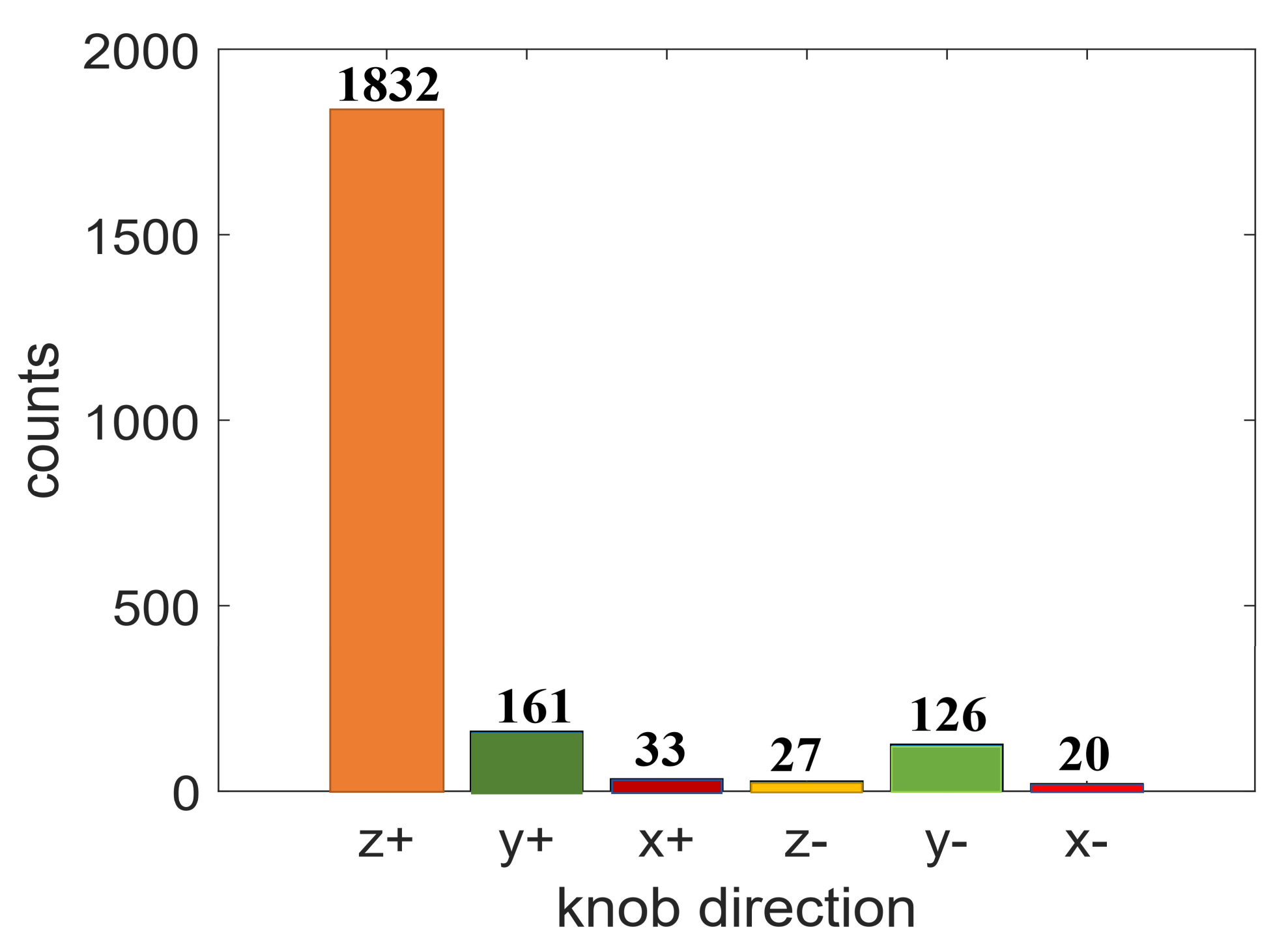}
  \includegraphics[width=0.52\columnwidth]{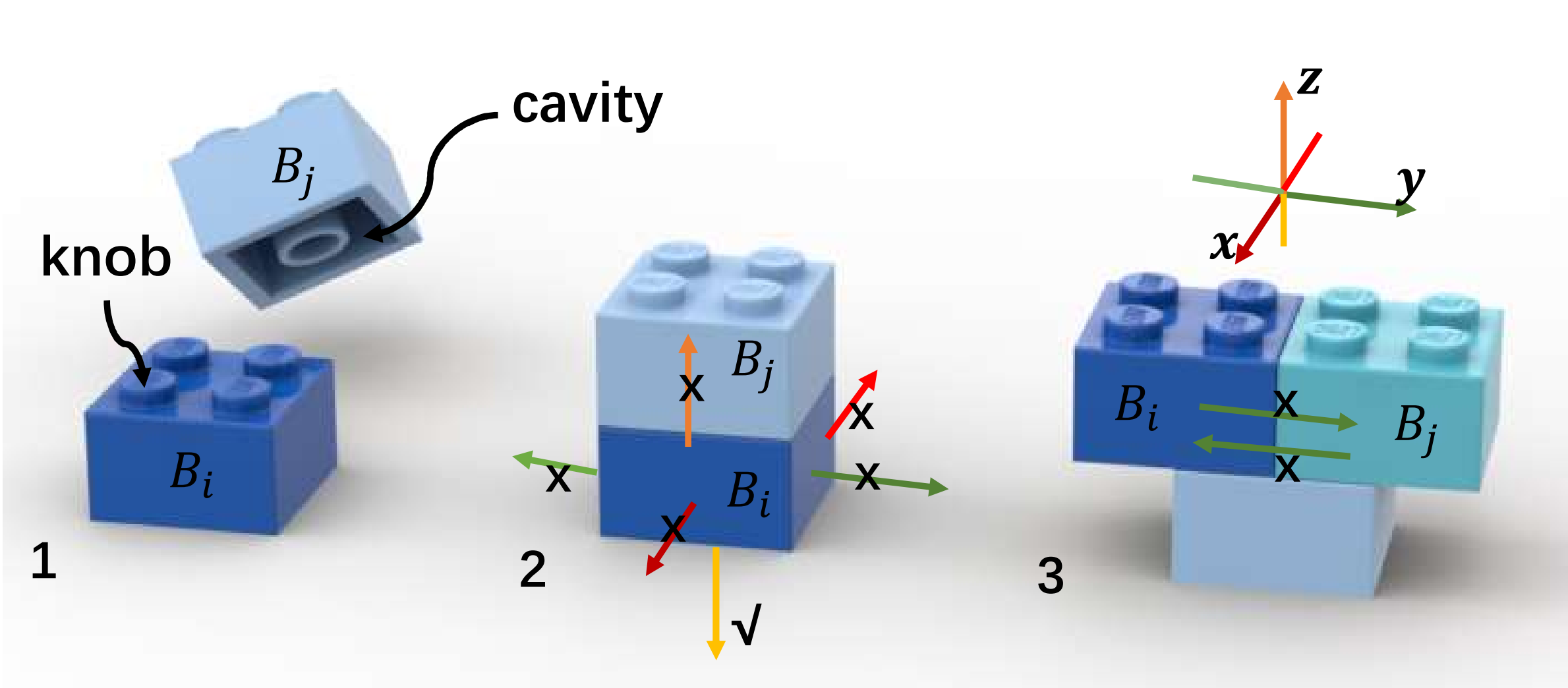}
  \caption{Statistics on knob directions.}
  \label{fig:knob}
\end{figure}

What should be clarified carefully id the mode of connection. As it is shown in Fig. \ref{fig:knob}.1, two blocks are snapped together by a knob from one brick and a cavity from another, which ensures the stability of the LEGO models.

\subsection{Heterogeneous graph}
It is intuitive to regard the bricks as nodes and their relationships as edges. In this way, we convert LEGO models into graph structures.  Nevertheless, there is additional information about the nodes and edges. In the next part, we will illustrate the node features and edge features extracted from the original LEGO model in detail.

\textbf{1) Node features}

In the LEGO model, every brick has a unique feature, so regarding bricks as identical nodes will cause information loss. Therefore, we need to categorize nodes to different types and capture the unique feature of nodes. As mentioned above, we labeled 23 types of frequently used bricks and the rest of the bricks are equally labeled as uncommon-type. In this method, the node feature corresponding to types is represented as a one-hot vector of 24 dimensions.

In addition, we also consider the positions as important features. Nevertheless, the pattern of positions is not consistent in different LEGO models for the following reasons. First, the start positions and views are randomly set when customers create their models, so the entire models are casually scattered. Second, users often rotate the models for convenience or a better perspective. Given the inconsistency of positions, we deliberately construct a coordinate system to make the different LEGO models as consistent as possible. In summary, there are three steps to determine the positions of bricks.

\begin{itemize}

\item The first step is to determine the direction of the coordinate system. By observing many LEGO assembly videos and parsing LEGO product instructions, we discovered a \textit{principle assembly direction}. This direction is consistent with the direction in which the majority of bricks' knobs orient, so we define it as the positive direction of the $z$-axis (vertical). Analogously, the second assembly direction is the direction in which bricks' knobs orient only ranks second to the principal assembly direction. Then the second assembly direction is defined as the positive direction of the $y$-axis (transversal). As the positive directions of the $y$-axis and $z$-axis  are fixed, we can easily infer the positive direction of the $x$-axis (longitude) in a left-half coordinate system. Moreover, we also verify the rationality of this setting by exploring the structures of the collected LEGO models. As shown in Fig. \ref{fig:knob}, nearly all the knobs are oriented in the principle assembly direction, which is set to be positive direction of the $z$-axis ($z$+ for short, and the same manner with $y+, y-, x+, x-$), and very few knobs are oriented in $z$-.  Moreover, the number of knobs orient $y$+ is close to the number of knobs orient $y$-, which may be caused by symmetric structures.

\item The second step is to determine the origin  of the coordinate system $\mathscr{O}$. We find the bounding box of the whole LEGO model, denoted as $\mathscr{B}_m$. The origin is set to be the vertex on $\mathscr{B}_m$, whose $x$, $y$, $z$ coordinate is the smallest.

\item Considering the coordinate system is constructed, the third step is to calculate the final positions of the nodes. For $i$-th node $v_i$ corresponding to brick $B_i$.  We denote its bounding box as  $\mathscr{B}_i, i=1, ..., n$, where $n$ represents the number of bricks in the LEGO model.  For $v_i$, we set its position as the center of the $\mathscr{B}_i$. After we got all the positions of the nodes, we normalize them to the unit cube as the final node features of positions.

\end{itemize}

\textbf{2) Edge features}

In graph structure, edges between nodes represent the relationships between bricks. In the LEGO model, the most obvious relationship between bricks is link.  However, the block relationship between nodes is equally important, which determines whether the assembly process is feasible. Next, we will illustrate the two types of relationships and their corresponding edges in detail.
\begin{figure*}
\centering
  \includegraphics[width=2\columnwidth]{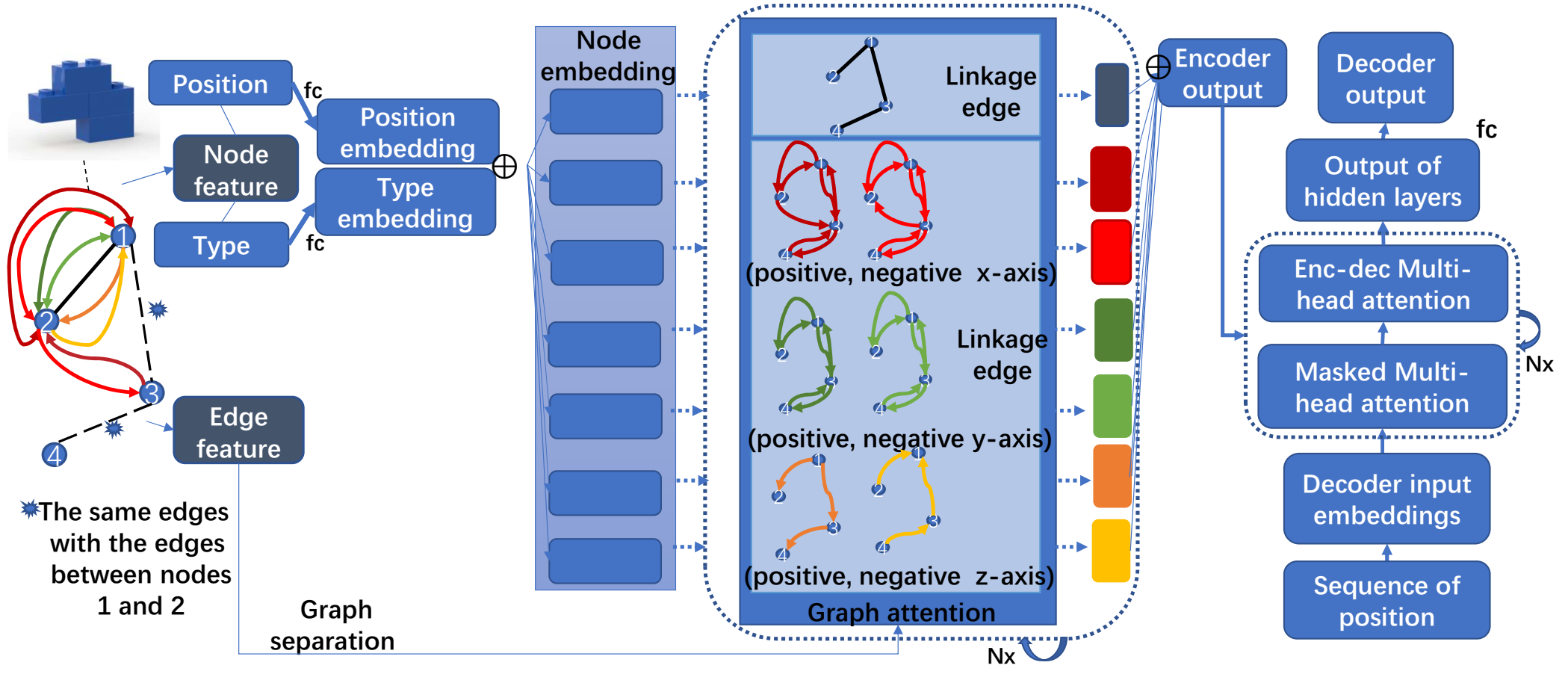}
  \caption{Graph-transformer framework}
  \label{fig:HGAT}
\end{figure*}
\begin{itemize}

\item Link edge: if bricks $B_i$ and $B_j$ are connected with a knob-cavity structure. Then there is a link edge between the corresponding nodes $v_i$ and $v_j$. This edge is called a link edge, denoted as relation triplet $⟨v_i, links, v_j⟩$, and $⟨v_j, links, v_i⟩$.  As shown in fig. \ref{fig:knob}.2.

\item Block edge: Compared with the link relationship, the block relationship is more intricate. If brick $B_i$ collides with brick $B_j$ when it moves a short distance along the direction of the axes ( both positive and negative), namely, brick $B_i$ cannot be assembled or removed under the condition that brick $B_j$ is already assembled. Then there is a directed edge from the corresponding node $v_j$ to $v_i$. In this way, the block relationship has directions. We treat different block directions as different edges. It is also denoted as a relation triplet, for example, $⟨v_i, block \ in \ directions \ of \ the \ positive \ x-axis, v_j ⟩$. As shown in Fig. \ref{fig:knob}.2, brick $B_i$ can only be snapped together with $B_j$ along with the direction of the positive $z$-axis.  It is blocked by $v_j$ in the direction of the $x$-axis (positive, negative), $y$-axis (positive, negative), and $z$-axis (negative). What should be clarified is that the block relationship exists not only between linked bricks but also between adjacent bricks. As shown in \ref{fig:knob}.3, $B_i$ and $B_j$ are placed side-by-side and blocked with each other in the positive and negative directions of the $y$-axis.

\end{itemize}

So far, we have constructed edge-heterogeneous graph structures.  As mentioned in part A, we have fixed all the separations, that is, all the bricks are connected with the knob-cavity structure, which guarantees the connectivity of the generated graph structure.

\subsection{ Ground Truth}
In the LEGO Studio platform, users can freely create their characters. They can causally add or remove the bricks even if other bricks are blocking them. However, it is infeasible in physical assembly. Therefore, we cannot use this creation sequence as the ground truth for assembly, even though the LEGO Studio has preserved it.
In order to efficiently obtain a feasible and reasonable assembly sequence as ground truth, we adopt a two-step process.

\begin{itemize}

\item First, we generate a feasible assembly sequence in the assembly-by-disassembly approach proposed in \cite{zhou2022brick}. However, the pattern of our self-collected LEGO models is different from the humanoid LEGO models, so the generated assembly sequence is not as good as it is in \cite{zhou2022brick}. Therefore, there is a additional modification step.

\item Secondly, considering the reasonability of the assembly sequence, we manually adapt the assembly sequence according to the following principles. (1) The order of substructures with similar semantic meaning are sequential. (2) For spatial continuity, we modified the situation if two bricks were located far away but assembled one after another.

\end{itemize}

\section{Graph-transformer for ASP problem}

Considering the particularity of the ASP problem in the LEGO model, we propose a new graph-transformer framework to predict an assembly sequence. The graph-transformer framework is an end-to-end framework for which the input is a heterogeneous graph and the output is the predicted assembly positions.

In Section III, we abstract the LEGO model to the graph structure, so the following statement will be based on the terminology of the graph structure. Our framework consists of two parts, graph encoder and transformer decoder. The work flow of our framework is shown in Fig. \ref{fig:HGAT}. For conciseness, the normalization layer is left out and will also not be mentioned in the following model description.

\subsection{Heterogeneous Graph encoder}
\subsubsection{Graph attention}
The majority of transformer-type models adopt the tuple $(Q, K, V)$ to perform the attention mechanism, represented in the following expression:\\
$$ Attention(Q, K, V)= softmax(\frac{QK^T}{\sqrt{d_k}})V$$
We degenerate this expression to node level, for the $i$-th node:\\
$$ Attention(Q_i, K,V)=\Sigma_{j=1}^n softmax(\frac{Q_iK_j}{\sqrt{d_k}}) V_j $$
In the above expression, $n$ is the number of tokens. Transferring to our setup, $n$ is the number of nodes in the graph structure. This means all of the nodes in the graph structure have an impact on the $i$-th node and they are treated equally. To this end, the graph structure has not been well explored and utilized, for which we adopt the treatment in \cite{velivckovic2017graph}. We perform masked attention— only consider the $i$-th node for $j$-th node, $i \in \mathscr{N}_i$  where $\mathscr{N}_i$ is neighbor set of $i$ (connect $i$-th node with edges).   In this way, the attention of $i$-th node becomes:
$$ Attention(Q_i, K,V)=\Sigma_{j \in \mathscr{N}_i} softmax(\frac{Q_iK_j}{\sqrt{d_k}}) V_j $$

\subsubsection{Heterogeneous edges for attention mechanism}

In above part, we restricted the attention mechanism to the neighbors, which preserves the graph structure to a certain extent. However, it does not take into account the types of edge. As mentioned in section III, there are link edges and block edges. Meanwhile, the block edges have different directions. In the assembly process, they are of great difference, so we discriminate them in the graph-transformer framework. We no longer use shared linear transformations for $Q, K, V$ but represent each type of edge with an individual linear transformation:
$$Attention(Q_i,K,V)=\Sigma_{t=1}^7 \Sigma_{j \in \mathscr{N}_i^t} softmax(\frac{Q_i^t K_j^t}{\sqrt{d_k}}) V_j^t$$
where the $t=1,2,..,7$ represent different edge types, that is, link edge and six types of block edge;  $\mathscr{N}_i^t$ represents the set of neighbors connected with $t$-th edge type and $Q^t, K^t, V^t$ are
obtained under the linear transformation corresponding to $t$-th edge type.

As it is shown in Fig. \ref{fig:HGAT}, this heterogeneous graph attention is equivalent to dismembering the heterogeneous graph into homogeneous “subgraphs” according to the types of edges. After applying graph attention independently, we aggregate the attention of all "subgraphs". Iterating this process in a stack of N identical layers, we got the final encoding output.

\subsection{Transformer decoder}

As shown in Fig. \ref{fig:HGAT}, the decoder part of our model is identical to the classical transformer decoder, so we will not illustrate it in detail. Nevertheless, we would like to clarify that the input and output of our decoder are sequences of position, that is, our target is to predict the next predicted position. In this way, we set the next node (or brick to be assembled) to be the one whose distance to the predicted location is the shortest.  To this end, we set the loss function to be the Mean Squared Error (MSE) between the real sequence of positions and the predicted sequence of positions.

\section{Experiment}
This section consists of four parts. First, we state the metrics used to evaluate the similarity of the predicted and the ground truth. Second, we demonstrate the impact of anomalous data items. Third, we test the effect of node types and edge types in ablation studies. Finally, we provide qualitative and quantitative results of our model on a in-process task.

\subsection{Metrics}

The quality of the predicted assembly sequence is evaluated using two metrics: Kendall's $\tau$ and the Regularized Location Swap Deviation (RLSD). The following Kendall's $\tau$ is used to measure the similarity of node priorities.
$$ \tau = \frac{N_{concordant}-N_{disconcordant}}{n(n-2)/2}$$
Where $n$ represents the length of the sequence and the $N_{concordant}$ and $N_{disconcordant}$ respectively represent the number of concordant pairs and disconcordant pairs. We would like to demonstrate the concordant pairs in our setup with the following example: if brick $B_i$ is assembled earlier (or later) than brick $B_j$ in both ground truth and predicted sequence, then we call it a concordant pair, and call it disconcordant pair for the opposing situation.

The second metric is Regularized Location Square Deviation. The conventional LSD can be viewed as the mean squared error (MSE) for components’ locations, $i[A],i[B]$.
$$LSD(A,B)=\frac{1}{n} \Sigma_{i=1}^{n-1} (i[A]-i[B])^2$$
However, the value of LSD greatly depends on the length of a sequence, thus it is divided by a coefficient of regularization $(n+1)(n-1)/2$. In this way, the value of RLSD is constrained in $[0, 1]$. The closer this value is to 0, the better the predicted sequence.

\subsection{Impact of large graphs}
According to Fig. \ref{fig:LEGO_model} in Section II, there is a large difference in the number of bricks in an individual model, which means the number of nodes in a graph varies. The more nodes a graph has, the more complex its structure is. And it will tends to have a hierarchical structure, so its pattern will no longer be consistent with short ones and it will be an outlier with a high probability. To this end, we eliminate top $20\%$ largest graphs and $40\%$ largest graphs from the dataset, The result is shown in TABLE \ref{table:1}.

\begin{table}[h]
\caption{Impact of large graphs}
\label{table:1}
\begin{center}

\begin{tabular}{|c|c|c|c|}
\hline
     & whole data & smallest $80\%$  & smallest $60\%$\\
\hline
Kendall's  $\tau$ $\uparrow$ & 0.1700 & 0.2267 & \textbf{0.4482}  \\
\hline
RLSD        $\downarrow$      & 0.3948  & 0.3711 & \textbf{0.2440}  \\
\hline
\end{tabular}
\end{center}

\end{table}

The smallest $80\%$ means we use only $80\%$ of data items whose number of nodes is smallest, in which $80\%*75\%$ is training set and the rest $80\%*25\%$ is test set. The same is true for the smallest $60\%$. The result in the next parts is also obtained under the setting: $ size \ of \ training \ set: size \  of \ test \   set=3:1$

The results show that when the number of nodes is small, the pattern of the graphs is more consistent. Under this condition, the graph-transformer framework is able to capture general rules. Kendall's $\tau$ is around 0.44, which is not a strong correlation but a moderate one. However, we still regard it as meaningful for the following reasons. First, the Kendall's $\tau$ is larger than 0 significantly, which means our predicted sequence has positive correlation with ground truth, so it captures the trend of assembly to certain degree. Second, in this method, we can obtain a sequence whose similarity to a feasible solution is 44 percent. Based on this sequence, we drastically reduce the search space of an NP-complete problem. Moreover, the RLSD is around 0.24, meaning that transferring this sequence to the ground truth requires much less swapping of elements in the sequence. Finally, we are aiming at generate a benchmark for future research.

\subsection{Ablation study}
In this part, a series of ablation studies are conducted to show the effectiveness of different graph features. The results are shown in TABLE \ref{table:ablation}. First, we explore the node features. Since the output of our graph-transformer framework is a sequence of positions, thus, as features of nodes, the position is unelimatable for the model. To this end, we only explore the effect of node type. Unexpectedly, rather than improving the performance, node type has a negative impact on the performance. After careful consideration, we find a possible reason to account for this situation.
It is the random combination of different types of bricks that ensures that a variety of 3D shapes can be generated. Consequently, the type of nodes does not have an effect in predicting the order of the assembly, and acts as a piecemeal information during training.
Moreover, we explore the effect of edge type.  The results show that there has been a huge improvement when we add both the link and block edge to the graph-transformer framework. This result implies that both link and block edges are critical factors for the ASP problem, and they play different roles.

\begin{table}[h]
\begin{center}
\caption{Ablation study}

\label{table:ablation}
\begin{tabular}{|l|ll|ll|}
\hline
                   & \multicolumn{2}{l|}{position}                                & \multicolumn{2}{l|}{position \& type} \\ \cline{2-5}
\multirow{-2}{*}{} & \multicolumn{1}{l|}{$\tau$ $\uparrow$} & RLSD  $\downarrow$                          & \multicolumn{1}{l|}{$\tau$ $\uparrow$}  & RLSD $\downarrow$   \\ \hline
Link edge          & \multicolumn{1}{l|}{0.3717} & 0.2892                         & \multicolumn{1}{l|}{0.2764}  & 0.3240 \\ \hline
Block edge         & \multicolumn{1}{l|}{0.3630} & 0.2944                         & \multicolumn{1}{l|}{0.3338}  & 0.2954 \\ \hline
Link \& Block      & \multicolumn{1}{l|}{0.4482} & 0.2440 & \multicolumn{1}{l|}{0.4325}  & 0.2422 \\ \hline
\end{tabular}%

\end{center}
\end{table}

\subsection{In-process task}
\begin{figure}
\centering
  \includegraphics[width=0.45\columnwidth]{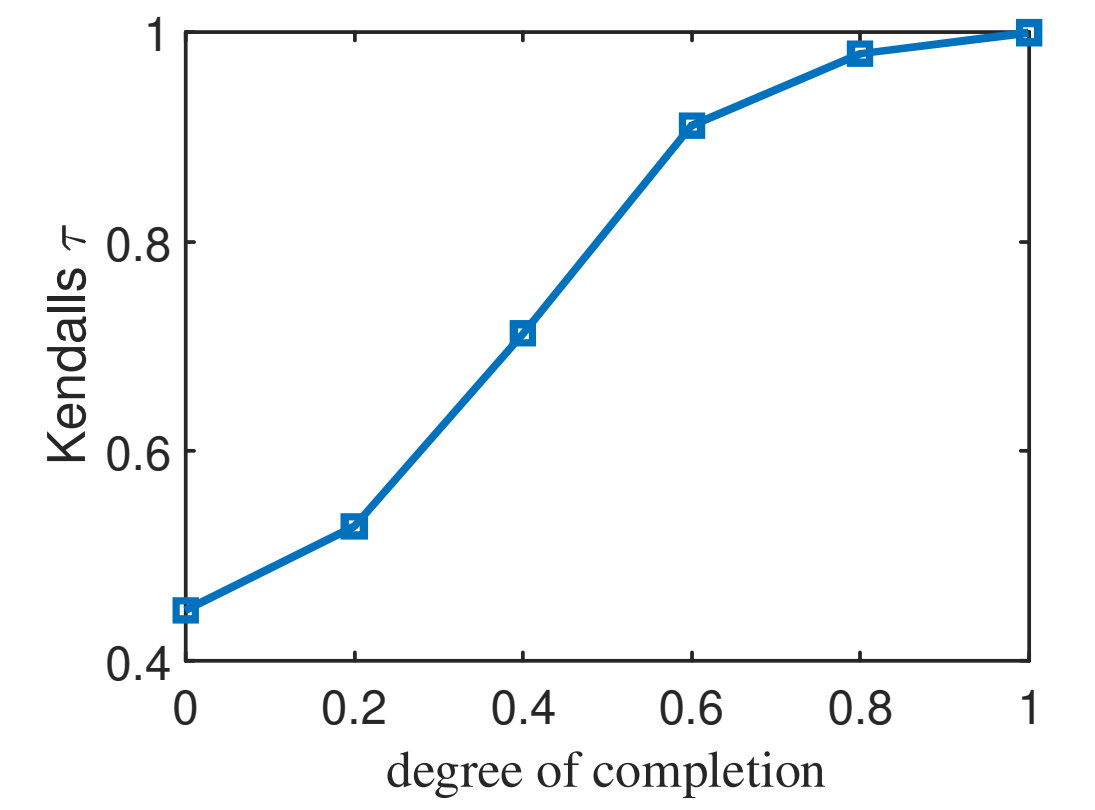}
  \includegraphics[width=0.45\columnwidth]{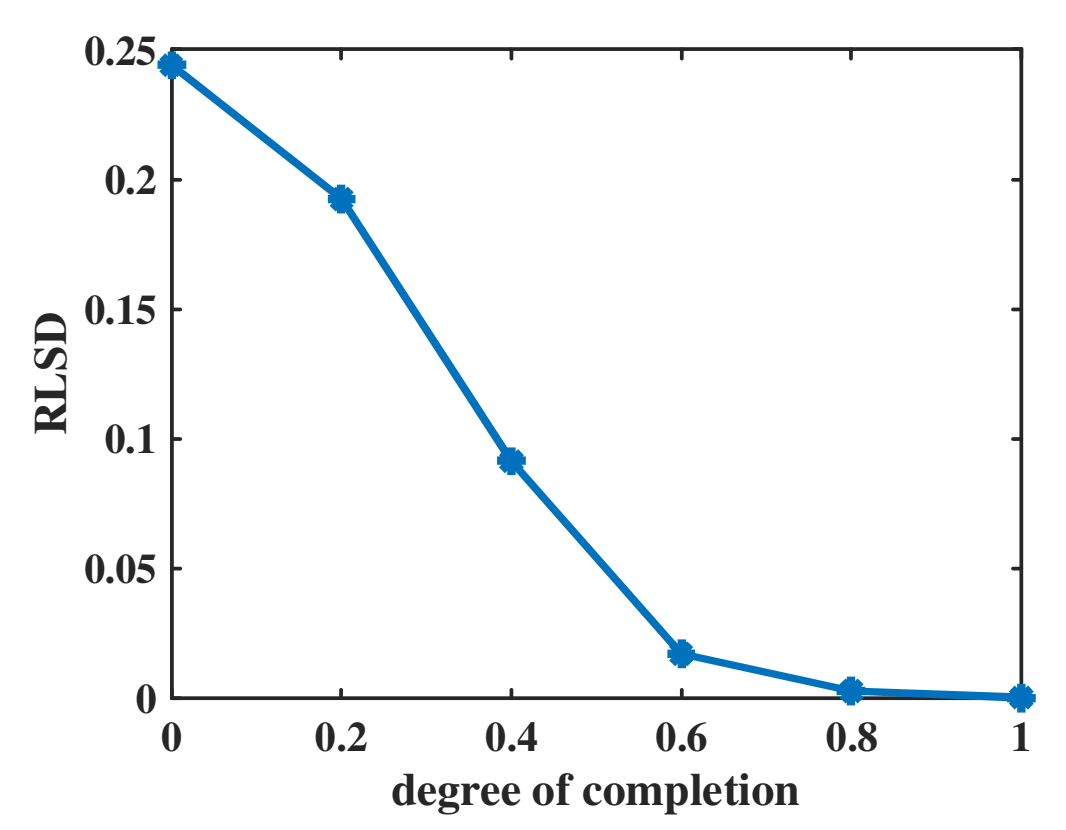}
  \caption{Result of in-process task.}
  \label{fig:in-process-accu}
\end{figure}

In this part, we design an in-process assembly sequence prediction task. Assuming that a LEGO model is partially assembled, we test the performance of our graph-transformer framwork by completing the following assembly procedure. In detail, we set the degrees of completion as $20\%$, $40\%$, $60\%$ and $80\%$. Based on these half-finished LEGO model, we complete the assembly sequence and obtain the results shown in Fig. \ref{fig:in-process-accu}.

\begin{figure}
\centering
  \includegraphics[width=0.95\columnwidth]{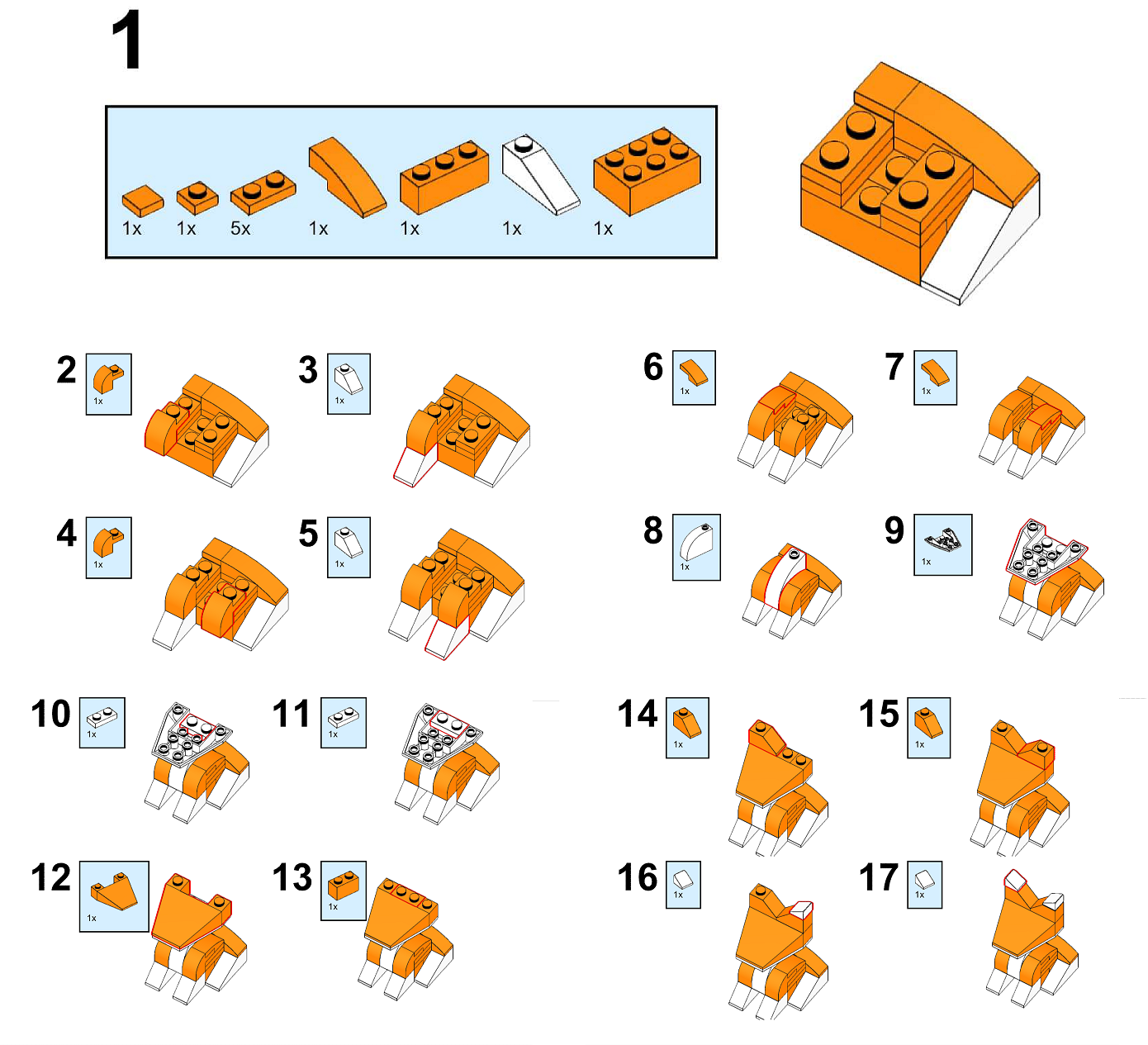}
  \caption{Result of in-process task.}
  \label{fig:in-process-example}
\end{figure}

As presented in Fig. \ref{fig:in-process-accu}, with the increase of the degree of completion, Kendall's $\tau$ increases and the RLSD drops, which means the accuracy of our prediction is increased rapidly.  Based on a given half-finished model, the graph-transformer framework has a preliminary understanding of the model and captures the unique features of the individual models. Consequently, it is a great advance over the results obtained by direct prediction in part B. Meanwhile, we present an example of this in-process task in Fig. \ref{fig:in-process-example}. We initialize the model to be 40\%-completed and labeled it as 1 in the figure. The following results are feasible and reasonable.  In summary, our graph-transformer framework performs well on in-process tasks.

\section{Conclusion and future work }

In this paper, we present a self-collected dataset of LEGO models for the ASP problem and formulate it into a heterogeneous graph structure. In addition, we proposed a Graph-transformer model for heterogeneous graph to predict an assembly sequence. However, our model does not perform well on large graphs. In the future, we would like to organize the nodes in hierarchical structure. In this way, the task for large graphs can be divided into hierarchical subtasks, which can be solved efficiently and efficiently by our Graph-transformer framework.

\addtolength{\textheight}{-6cm}   

\bibliographystyle{IEEEtran}
\bibliography{IEEEfull,root}

\end{document}